\documentclass[final,3p,times]{elsarticle}

\usepackage{amssymb}
\usepackage{amsmath}

\usepackage{colortbl}
\usepackage{xcolor}
\usepackage{url}
\usepackage{multirow}
\usepackage{makecell}

\journal{NeuroComputing}

\begin{document}

\begin{frontmatter}



\title{FCN+$\colon$ Global Receptive Convolution Makes FCN Great Again}


\author[label1]{Xiaoyu Ren} 
\ead{renxiaoyu@mail.iap.ac.cn}
\author[label2]{Zhongying Deng\corref{Corresponding}}
\cortext[Corresponding]{Corresponding author}
\ead{zd294@cam.ac.uk}
\author[label3]{Jin Ye}
\author[label3]{Junjun He}
\author[label1]{Dongxu Yang}
\ead{yangdx@mail.iap.ac.cn}
\affiliation[label1]{organization={Institute of Atmospheric Physics, Chinese Academy of Sciences},
            city={Beijing},
            postcode={100029}, 
            country={China}}
\affiliation[label2]{organization={University of Cambridge},
            city={Cambridge},
            postcode={CB30WA}, 
            country={United Kingdom}}
\affiliation[label3]{organization={Shanghai Artificial Intelligence Laboratory},
            city={Shanghai},
            postcode={200232}, 
            country={China}
            }

\begin{abstract}
Fully convolutional network (FCN) is a seminal work for semantic segmentation. However, due to its limited receptive field, FCN cannot effectively capture global context information which is vital for semantic segmentation. As a result, it is beaten by state-of-the-art methods which leverage different filter sizes for larger receptive fields.  
  In this paper, we propose a novel global receptive convolution (GRC) to effectively increase the receptive field of FCN for context information extraction, which results in an improved FCN termed FCN+. The GRC provides \emph{global receptive field} for convolution \emph{without introducing any extra learnable parameters}.  
  Specifically, the GRC first divides the channels of the filter into two groups. The grid sampling locations of the first group are shifted to different spatial coordinates across the whole feature map, according to their channel indexes. This can help the convolutional filter capture the global context information. The grid sampling location of the second group remains unchanged to keep the original location information.
  Convolving using these two groups, the GRC can integrate the global context into the original location information of each pixel for better dense prediction results. With the GRC built in, FCN+ can achieve comparable performance to state-of-the-art methods for semantic segmentation tasks, as verified on PASCAL VOC 2012, Cityscapes and ADE20K. {Our code will be released at \url{https://github.com/Zhongying-Deng/FCN_Plus}.}
\end{abstract}


\begin{highlights}
\item We propose a global receptive convolution (GRC) to provide a global receptive field for convolutional filters without introducing any extra learnable parameters. 
\item Based on GRC and Fully Connected Network (FCN), we build a novel semantic segmentation network, termed FCN+. 
\item FCN+ can achieve comparable performance to state-of-the-art methods on semantic segmentation datasets like PASCAL VOC 2012, Cityscapes, and ADE20K.
\end{highlights}


\end{frontmatter}




\section{Introduction}

Semantic segmentation is a dense prediction task which aims to assign each pixel a class label. It is a fundamental yet challenging task in computer vision, with a variety of applications in scene understanding ~\cite{zhao2017pyramid,zhao2018psanet,zhang2024segcft}, autonomous vehicles~\cite{yang2018denseaspp,yang2024msvfe} and medical image diagnostics~\cite{ronneberger2015u,ye2022gid,ge2023unsupervised}. 
The challenges mainly lie in the large-scale variation of objects and stuff, as well as the similar appearance of different objects. Concretely, the \textbf{challenge one} is that: extremely large-scale objects may cause inconsistent segmentation results on the same object while extremely small-scale objects can possibly be neglected. To alleviate this issue, multi-scale feature representations 
are heavily used to capture objects of different scales. 
The \textbf{challenge two}: similar visual appearance of different objects probably leads to confusion on the boundaries of them. To better segment these boundaries, it is necessary to integrate global context information into each pixel's representation so that overall scene understanding can be employed for correctly classifying each pixel on the boundaries.

Fully convolutional network (FCN)~\cite{long2015fully} is a seminal work for semantic segmentation. It leverages deep convolutional neural networks (CNNs) to extract features for objects and scenes, and further aggregates feature maps of different spatial resolutions from hierarchical layers as multi-scale representations. Owing to this, FCN is robust to objects' scale variance to a certain extent. However, due to its limited receptive field, FCN can hardly capture \emph{global} context information. 

\begin{figure}
    \centering
    \includegraphics[trim=70 10 45 10,clip,width=0.95\textwidth]{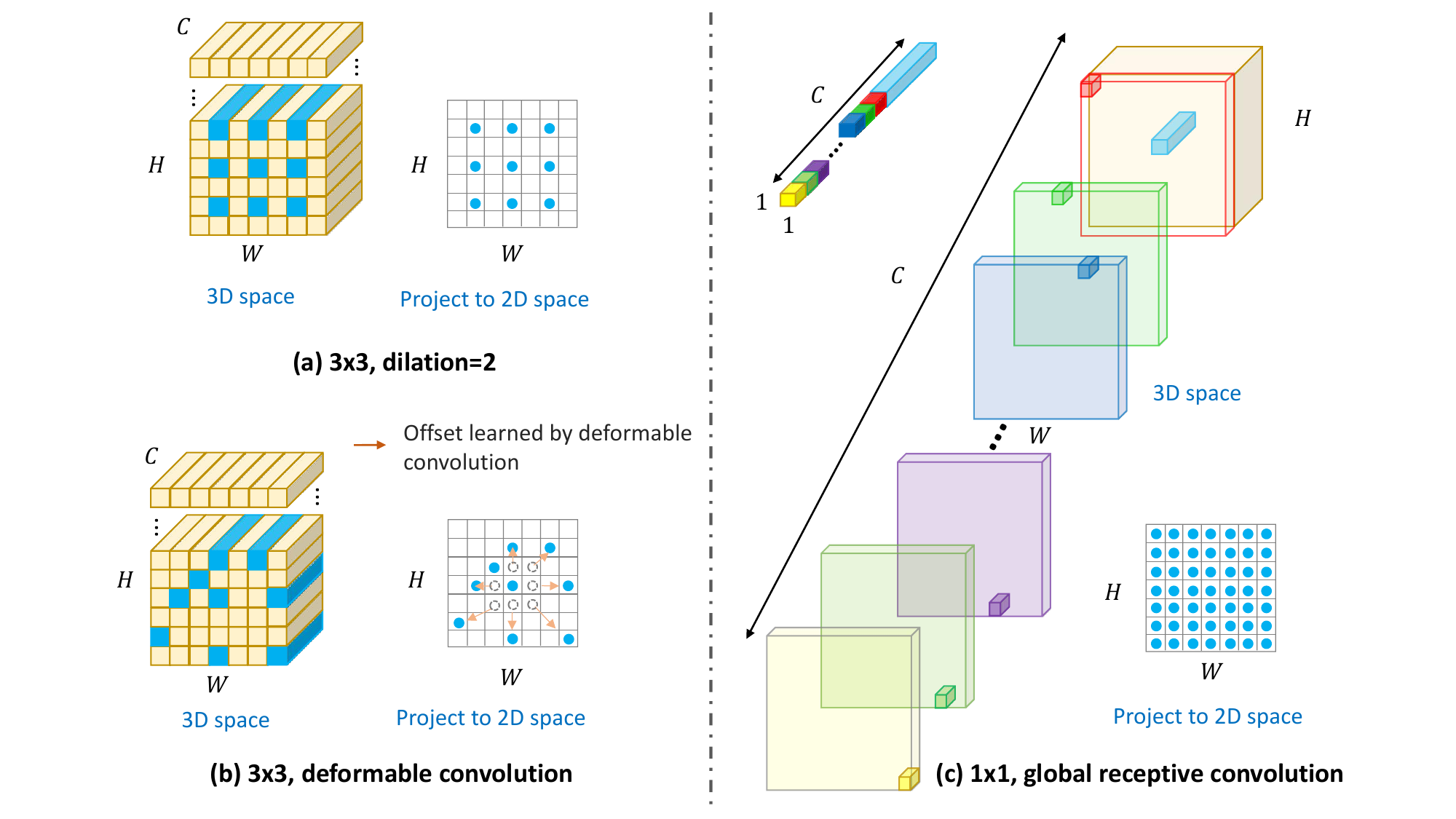}
    \vspace{-0.2cm}
    \caption{Comparison of global receptive convolution with the other convolutions. (a) 3$\times$3 dilated convolution with dilation=2. (b) 3$\times$3 deformable convolution. Deformable convolution learns offset for each spatial location to enlarge receptive fields. For both (a) and (b), different channels of their filters share exactly the same grid sampling location, thus having limited receptive fields. (c) 1$\times$1 global receptive convolution (GRC). The GRC divides the 1$\times$1 kernel into two groups along the channel dimension, e.g., the light blue and the other colors. The group of light blue convolves the central pixel to keep the original location information. While the group of other colors are shifted to other spatial locations to capture the global context information. Note that 1$\times$1 GRC has a global receptive field, much larger than that of 3$\times$3 dilated convolution or deformable convolution.}
    \label{fig:intro_cmp_grc}
    \vspace{-0.3cm}
\end{figure}

Other methods try to address these challenges by introducing pooling layers of different pooling grids, or convolutions of different dilation rates or different grid sampling locations~\cite{zhao2017pyramid,chen2014semantic,chen2017deeplab,chen2017rethinking,chen2018encoder,dai2017deformable}. Nevertheless, pooling layers can lose finer spatial information~\cite{zhao2017pyramid}, leading to poor representations for some pixels; large dilation rates may cause gridding artifacts~\cite{chen2017deeplab,chen2017rethinking} and lose neighbor information; learning grid sampling locations increases the model parameters~\cite{dai2017deformable} and the learned locations can be inexact. More importantly, all of these methods assume that different channels of a filter share exactly the same grid sampling locations which are limited to fixed local spatial coordinates, as shown in Fig.~\ref{fig:intro_cmp_grc}(a) and (b). 
For example, the grid sampling location of a 1$\times$1 convolution is a fixed spatial point at the feature map---no matter how many channels are involved. This is infeasible because it can never enlarge the receptive field of 1$\times$1 convolution. 
These limitations hinder their efficacy or efficiency for integrating \emph{global} context information to pixel-wise representations.

To remedy the above limitations, we propose a global receptive convolution (GRC). The motivation is that different channels of a filter can have different grid sampling locations, which can enlarge the receptive field. Based on this simple idea, the GRC is designed to provide \emph{global receptive field} for convolutional filters \emph{without introducing any extra learnable parameters}.  In Fig.~\ref{fig:intro_cmp_grc}, we apply the GRC to  1$\times$1 convolution for illustration.  
Specifically, the GRC first divides the channels of filters into two groups. Then it shifts the grid sampling locations of the first group to different coordinates (across the whole input image) according to the channel index, so that the filter can capture the global context information. The grid sampling locations of the second group remain unchanged to keep the original location information.  Thus, the GRC can integrate the global context into the original location information for each pixel, further tackling the \textbf{challenge two} for better dense prediction results. The GRC is simple yet effective, and can be flexibly applied to any other filter sizes.

We then apply the GRC to the feature extractor of FCN by replacing the standard convolution with our GRC. This results in a new architecture which we term FCN+. Here, FCN is chosen for two reasons: 1) FCN is one of the most popular methods in semantic segmentation which is used as a baseline network for many state-of-the-art methods~\cite{badrinarayanan2017segnet,zhao2017pyramid,chaurasia2017linknet}. These methods use FCN-based encoder which is transferred from classification models pre-trained on large-scale dataset like ImageNet~\cite{russakovsky2015imagenet}, but they also design complicated decoders, such as PPM (pyramid pooling module)~\cite{zhao2017pyramid} and ASPP (atrous spatial pyramid pooling)~\cite{chen2017deeplab,chen2018encoder},  to better adapt the classification-based encoder models to semantic segmentation task. Complicated decoders are not efficient due to more parameters and FLOPs. Therefore, we provide another perspective to adapt the classification-based encoder model to segmentation tasks, which is to introduce the global receptive convolution (GRC) to the FCN-based encoder, without touching the decoders. We will show that with the GRC-based encoder, FCN+ can also beat these complicated decoder-based methods efficiently and effectively. 
2) FCN has already provided multi-scale representations, so it can alleviate the \textbf{challenge one}. Benefiting from the GRC and FCN, FCN+ can not only provide multi-scale representations, but also incorporate global context information to simultaneously address the aforementioned two challenges for semantic segmentation.

\textbf{Contributions.} We propose a global receptive convolution (GRC) to provide a global receptive field for convolutional filters without introducing any extra learnable parameters. Based on GRC and FCN, we build a novel semantic segmentation network, termed FCN+. FCN+ can achieve comparable performance to state-of-the-art methods on semantic segmentation datasets like PASCAL VOC 2012, Cityscapes, and ADE20K.

\section{Related Work}
Deep convolutional neural networks (CNNs) have achieved remarkable success in semantic segmentation~\cite{long2015fully,zhao2017pyramid,ronneberger2015u,chen2017deeplab,zhou2018unet++}. These methods usually adopt multi-scale feature representations and global context information to deal with the aforementioned two challenges.

\textbf{Multi-scale feature representations} are usually more robust to objects with large scale variation  (the challenge one). To achieve this, fully convolutional network (FCN)~\cite{long2015fully} aggregates feature maps of different spatial resolutions from hierarchical layers as multi-scale representations.
U-Net~\cite{ronneberger2015u} and U-Net++~\cite{zhou2018unet++} fuse multiple low-level feature maps extracted from an encoder with high-level features from a decoder.
PSP-Net~\cite{zhao2017pyramid} incorporates a pyramid pooling module, which exploits different pooling grids to aggregate multi-scale feature maps. 
Deeplab series~\cite{chen2017deeplab,chen2018encoder} adopts multiple convolution filters, each having different dilation rates, for multi-scale representations. 
Different from these artificially defined static networks, dynamic routing~\cite{li2020learning} is proposed to automatically generate computational routes according to the scale distribution of each image. 
{
~\cite{zhou2021deep} uses multi-scale representations for distorted image quality assessment which is a different task in our work of semantic segmentation. Instead of concentrating on multi-scale learning, our work pays attention to the convolution itself, with a goal of enlarging the receptive field of convolution layers for better segmentation performance. As such, we propose the novel global receptive convolution (GRC) and do not design new multi-scale feature fusion strategies.}

\textbf{Global context information} plays a vital role in addressing the challenge of boundaries. One of the simplest ways for obtaining global context is using the global average pooling, such as~\cite{liu2015parsenet,zhao2017pyramid}. However, this method cannot effectively integrate global context information into each pixel’s representation. To address this issue, attention is widely used. Self-attention is proven to be effective in capturing long-range dependencies~\cite{wang2018non,yin2020disentangled,fu2019dual,yuan2018ocnet} for global context. It can also integrate the global context to pixel-wise representation by query-key-based matrix multiplication. Recent works~\cite{chen2021transunet,zheng2021rethinking} further adopt attention-based transformer as backbone and favorably improve the segmentation results. However, the matrix multiplication in self-attention considerably increases computational complexity. 
{
Other latest methods either propose new regularization, like the physical structure regularization in ~\cite{liu2025physically}, or new modules, such as the boundary generation module and boundary-guided refinement module in ~\cite{yue2024boundary}, both of which contain multiple standard convolution operations.
}

Our proposed method differs from all the above methods in the following ways. Firstly, all the previous methods assume that the sampling locations are only dependent on spatial location regardless of channels. But in our GDC, they are conditioned on both spatial location and channel index. As a result, our GDC can even increase the receptive field for 1$\times$1 convolution, as in Fig.~\ref{fig:intro_cmp_grc}(c), but previous methods cannot. Secondly, limited by the sampling locations, previous CNN-based methods cannot provide global context information by using a single filter size. While our GDC can provide global context information even using a single filter size. This can result in better performance. Lastly, the CNN-based or attention-based methods usually introduce more parameters or FLOPs. In contrast, our GDC does not introduce any extra parameters, thus more efficient.

\section{Methodology}
Our FCN+ is based on fully convolutional network (FCN)~\cite{long2015fully}, with some standard convolutions of FCN replaced by the global receptive convolution (GRC). The GRC aims to integrate global context information to pixel-wise representation for better dense prediction results. The motivation for GRC is simple: 
different channels of a single filter can have different grid sampling locations. The channel-dependent sampling location can effectively enlarge the receptive field to the whole feature map without introducing any extra parameters, as depicted in Fig.~\ref{fig:intro_cmp_grc}. Below we first revisit standard convolution, dilated convolution, and deformable convolution as preliminary, then introduce the GRC in detail, finally present the FCN+.

\subsection{Preliminary}
The input of a convolution is a 3D feature map $F\in \mathbf{R}^{H\times W\times C}$, with $H, W, C$ representing height, width and channel depth. The convolution uses convolutional filters $W\in \mathbf{R}^{C\times K \times K \times C'}$, where $K$ is the filter size and $C'$ is the output channel depth, to convolve the input feature map. Here, we only consider the case of a single filter, i.e., $C'$=1, but it can be easily applied to multi-filters. Therefore, we will omit $C'$ below for simplicity. The output of the convolution, i.e., $F'\in \mathbf{R}^{H\times W\times C'}, C'=1$, is formulated as
\begin{equation}
\label{eq:general_conv}
    F'(h_0, w_0)=\sum_{c=0}^{C-1} \sum_{(h,w)\in \mathcal{N}} W(c, h, w) * F(h_0+h, w_0+w, c),
\end{equation}
where $F'(h_0, w_0)$ is output feature map at the spatial location $(h_0, w_0)$. The operator $*$ denotes convolution. $h, w$ are the coordinates of grid sampling locations, which are both constraint in a local neighbor area $\mathcal{N}$ centered on $(h_0, w_0)$. The location of $(h_0, w_0)$ is termed as central sampling location while the pixels at $(h_0, w_0)$ are called central pixels.

For standard convolution,  $(h, w)$ enumerates all the spatial locations in $\mathcal{N}$, where 
\begin{equation}
\label{eq:standard_conv}
     \mathcal{N}=\{(-r, -r), (-r, -r+1), ..., (r-1, r), (r, r)\},
\end{equation}
where $r=\lfloor \frac{K}{2} \rfloor$ is conditioned on the filter size $K$ while $K$ determines the receptive field. For dilated convolution~\cite{chen2017deeplab,chen2017rethinking}, $(h, w)$ are in
\begin{equation}
\label{eq:dilated_conv}
    \mathcal{N}=\{(-r, -r), (-r, -r+d), ..., (r-d, r), (r, r)\},
\end{equation}
with $r=\lfloor \frac{d(K-1)+1}{2} \rfloor$ and $d$ as dilation rate. For deformable convolution~\cite{dai2017deformable}, $(h, w)$ are learned from the input feature map $F$ by using a function $f_{\psi}$, with $\psi$ as learnable parameters. Fig.~\ref{fig:intro_cmp_grc}(a) and (b) illustrate the dilated convolution and deformable convolution, respectively.

For all these convolutions, a single filter only has limited receptive field. An intuitive way to enlarge the receptive field is to use multiple filters of varied sizes. However, this can either lose neighbor information or introduce more parameters, which further negatively affects their efficacy or efficiency for integrating \emph{global} context information to pixel-wise representations. These convolutions have limited receptive field mainly because their grid sampling locations are only conditioned on some local spatial locations regardless of channel index. Therefore, we try to enforce the grid sampling locations to condition on both spatial location and channel index for larger receptive field. 

\begin{figure*}
    \centering
    \includegraphics[trim=40 15 200 5,clip,width=0.9\textwidth]{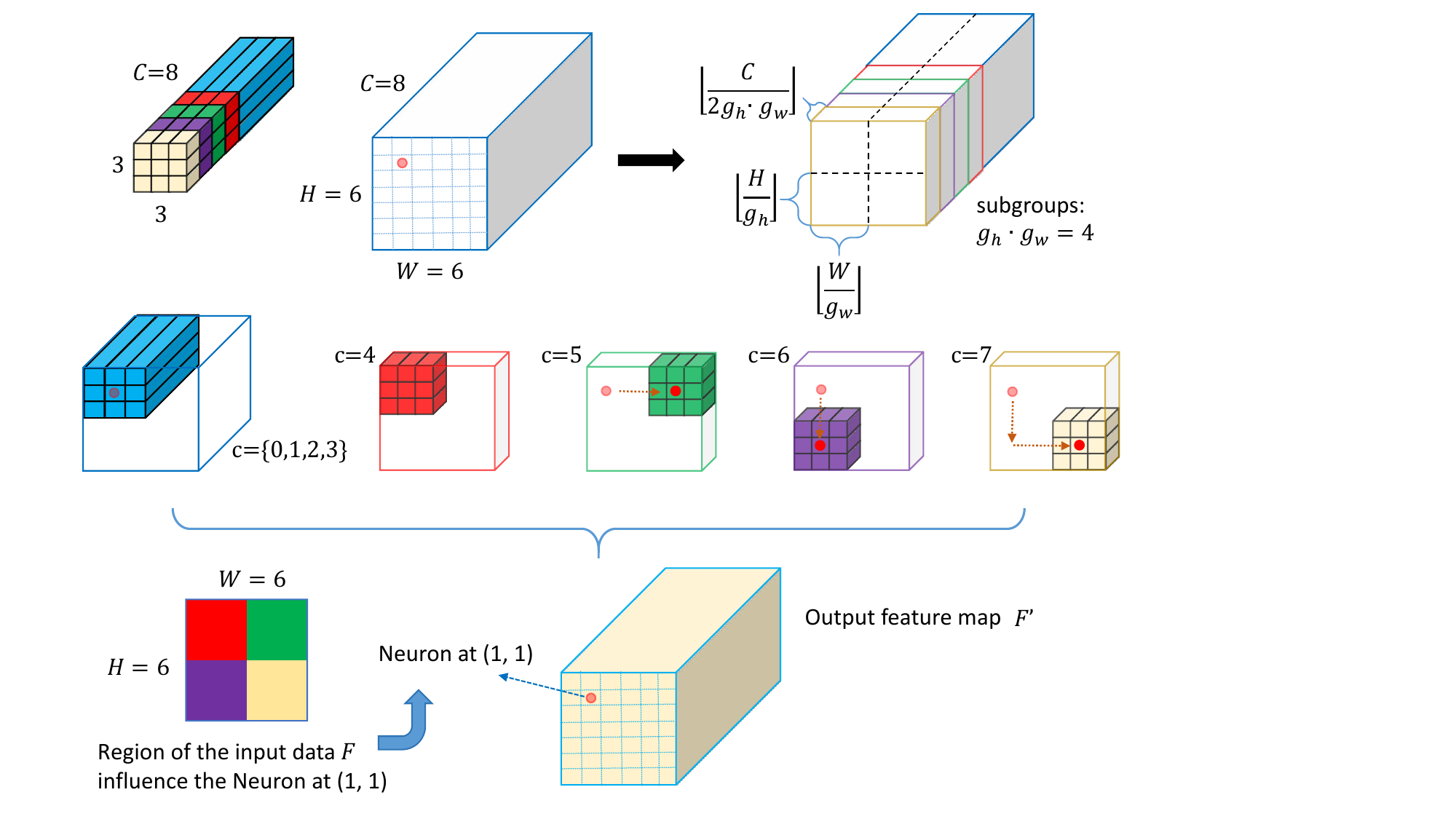}
    \caption{Illustration for 3$\times$3 global receptive convolution (GRC). This example shows how the GRC of 3$\times$3 filter size convolves a 6$\times$6$\times$8 input feature map $F$, with the central pixel $(h_0, w_0)=(1,1)$ (shown as the light red dot in the top row). GRC divides the channel into two groups. For the group of $c=\{0,1,2,3\}$, standard convolution is applied, as in the blue cube at the {middle} left. For the group of $c=\{4,5,6,7\}$, we further split it into $g_h \cdot g_w=4$ sub-group, with $g_h=g_w=2$. So each sub-group group has $\lfloor \frac{C}{2g_h\cdot g_w} \rfloor$=1 channel. Accordingly, the feature maps are divided into $g_h \cdot g_w=4$ sub-group, with spatial patch comprising $\frac{H}{g_h}\times\frac{W}{g_w}$ =3$\times$3 pixels. According to Eq. (6) or (7), $\hat{\mathcal{N}}=\{(0,0), (0, 3), (3,0), (3,3)\}$. Given channel index like $c=5$, we can obtain $n=1$ via Eq. (8). Thus, for the 5th channel ($c=5$), the offset is $\hat{\mathcal{N}}_1=(0,3)$. This offset, shown as the dashed arrow in the {middle} row, is then added to $(h_0, w_0)=(1,1)$ to shift the central sampling location. The new central sampling location is $(1,4)$. Centering on this location, the convolution is applied in the 5th channel-wise sub-group, as shown in the green cube. {Finally, the bottom row depicts the 2D region of the input data that influence the unit/neuron at the location $(1, 1)$ of the output feature map $F'$. It can be seen that the neuron $(1, 1)$ depends on the entire region of the input data, i.e., the red, green, purple, and yellow regions cover the whole 6$\times$6 input data.}
    }
    \label{fig:method_sampling_loc}
\end{figure*}

\subsection{Global Receptive Convolution}
We propose an effective and efficient convolution, dubbed global receptive convolution (GRC). The GRC provides \emph{global receptive field} for convolutional filters \emph{without introducing any extra learnable parameters}. The key to achieving this is to make different channels of a convolutional filter have different grid sampling locations across the whole input feature map. 
Specifically, we first divide the channels of the $W$ into two groups. Each group has equal channels by default, i.e., $\frac{C}{2}$. For the group of $c=\{0, ..., \frac{C}{2}-1\}$, we do standard convolution as in Eq.~\eqref{eq:general_conv} and ~\eqref{eq:standard_conv}. For the group of $c=\{\frac{C}{2}, ..., C-1\}$, we elaborately design its grid sampling locations, i.e., $\hat{\mathcal{N}}$,  for global receptive field. $\hat{\mathcal{N}}$ will be introduced later. Then, the GRC is formulated as
\begin{small}
\begin{equation}
\label{eq:grc}
    \begin{aligned}
    F'(h_0, w_0)&= \sum_{c=0}^{\frac{C}{2}-1} \sum_{(h,w)\in \mathcal{N}} W(c, h, w) * F(h_0+h, w_0+w, c) \\
    & +\sum_{c=\frac{C}{2}}^{C-1} \sum_{(\hat{h}, \hat{w})\in \hat{\mathcal{N}}} \sum_{(h, w)\in \mathcal{N}} W(c, h, w) *  F(h_0+\hat{h}+h, w_0+\hat{w}+w, c),
    \end{aligned}
\end{equation}
\end{small}
where $\mathcal{N}$ is the same as Eq.~\eqref{eq:standard_conv}. The first term is standard convolution as in Eq.~\eqref{eq:general_conv} except that the summation over channel $c$ is from 0 to $\frac{C}{2}-1$. This term keeps the original location information for the pixel-wise representation (i.e., for the feature representation of the central pixels at $(h_0, w_0)$).  In the second term, $\hat{\mathcal{N}}$ controls the central sampling location for GRC. The offsets $(\hat{h}, \hat{w})$ in $\hat{\mathcal{N}}$ shift the central location from $(h_0, w_0)$ to other locations according to the channel index $c$. In this way, the offsets can enable different grid sampling locations for capturing global context information.  
The design of the offsets will be detailed in the next paragraph. Finally, the global context information is integrated into pixel-wise representations by summing these two terms. See Fig.~\ref{fig:method_sampling_loc} for an illustration.

\begin{figure*}[t]
    \centering
    \includegraphics[trim=5 32 220 0,clip,width=\textwidth]{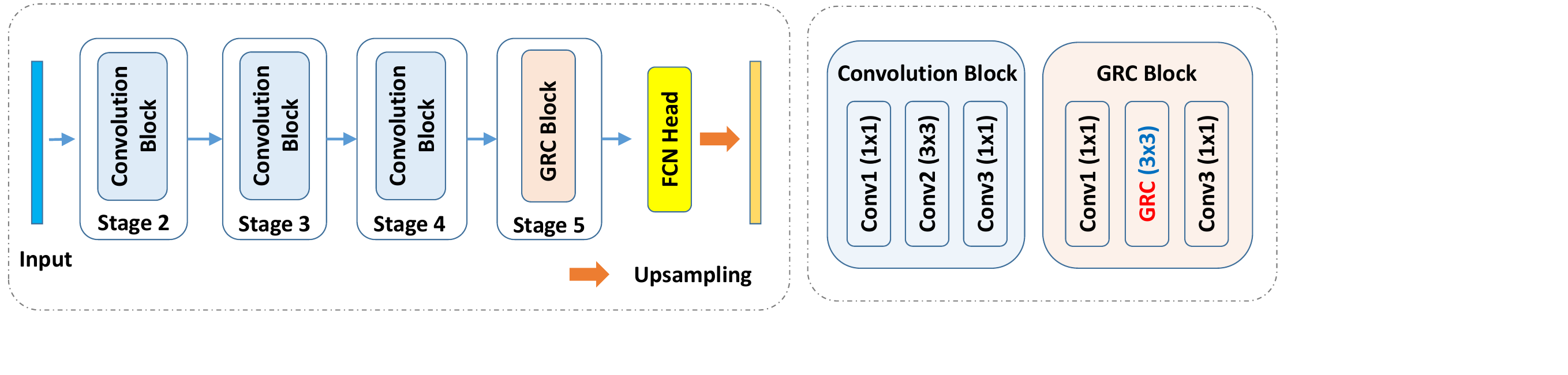}
    \caption{The architecture of FCN+. As in FCN~\cite{long2015fully}, FCN+ combines coarse high-level information with fine low-level information. FCN+ adopts a ResNet backbone which has five residual stages (Stage 1 omitted for simplicity). The implementation of FCN head can refer to MMSegmentation~\cite{mmseg2020}. The last stage of FCN+ is replaced with GRC-based residual blocks.
    }
    \label{fig:fcn_plus}
\end{figure*}

The offsets $\hat{h},\hat{w}$ control the shifted sampling locations. They are obtained by following steps. We further split the group of $c=\{\frac{C}{2}, ..., C-1\}$ into $g_h\cdot g_w$ sub-groups. So, each sub-group has $\lfloor \frac{C}{2g_h\cdot g_w} \rfloor$ channels. Correspondingly, we split the feature map into $g_h\cdot g_w$ patches along the spatial dimension, with each patch containing $\lfloor\frac{H}{g_h} \rfloor \cdot \lfloor\frac{W}{g_w} \rfloor$ pixels. Then the offsets are defined as
\begin{equation}
\label{eq:spatial_group}
    \hat{h}= i \cdot \lfloor\frac{H}{g_h} \rfloor, \quad \hat{w}= j\cdot \lfloor\frac{W}{g_w} \rfloor,
\end{equation}
where $i,j$ are patch indexes. They are integers that control the offsets, with $i\in[0, g_h-1], j\in[0, g_w-1]$. We will include the patch indexes $i,j$ to $\hat{h},\hat{w}$ for better clarity. This leads to $\hat{h}_i,\hat{w}_j$. 
Now we can obtain $\hat{\mathcal{N}}$ by
\begin{equation}
\label{eq:sampling_loc_ij}
    \hat{\mathcal{N}} = \cup_{i,j} \{(\hat{h}_i, \hat{w}_j)\}.
\end{equation}
More specifically, $\hat{\mathcal{N}}$ enumerates all the possible combinations of $i,j$, i.e.,
\begin{small}
\begin{equation}
\begin{aligned}
\label{eq:sampling_loc_enu}
    \hat{\mathcal{N}}=& \left \{ (0, 0), \left (0, \lfloor\frac{W}{g_w} \rfloor\right), ..., \left ((g_h-2)\cdot \lfloor\frac{H}{g_h} \rfloor, (g_w-1)\cdot \lfloor\frac{W}{g_w} \rfloor \right),   \right. \\ 
    & \left. \left( (g_h-1)\cdot \lfloor\frac{H}{g_h} \rfloor, (g_w-1)\cdot \lfloor\frac{W}{g_w} \rfloor \right) \right\}.
\end{aligned}
\end{equation}
\end{small}
$\hat{\mathcal{N}}$ comprises $g_h\cdot g_w$ offset coordinates in total. It almost covers the whole input feature map, i.e., $H\times W$, thus provides global receptive field for GRC. Recall that the key in GRC is to make different channels have different grid sampling locations. Hence, we choose different offset coordinates in $\hat{\mathcal{N}}$ according to the channel index $c$, or equivalently according to the channel-wise sub-group index $n$. Given $c$, $n$ can be formulated as
\begin{small}
\begin{equation}
\label{eq:sub_group_index}
    n = 
    \begin{cases}
      0, & \text{ if } \frac{C}{2}\le c  < \frac{C}{2}+\lfloor \frac{C}{2g_h\cdot g_w} \rfloor \\
      1, & \text{ if } \frac{C}{2}+\lfloor \frac{C}{2g_h\cdot g_w} \rfloor \le c  < \frac{C}{2}+2\cdot\lfloor \frac{C}{2g_h\cdot g_w} \rfloor .  \\
      ..., & ...\\
      (g_h\cdot g_w-1), & \text{ if } \frac{C}{2}+(g_h\cdot g_w-1)\lfloor \frac{C}{2g_h\cdot g_w} \rfloor \le c  < C  
    \end{cases}
\end{equation}
\end{small}
The $n$-th sub-group of channels corresponds to the $n$-th offset coordinates of $\hat{\mathcal{N}}$, denoted as $\hat{\mathcal{N}}_n$. Therefore, different channel groups have different offset coordinates, resulting in different shifted sampling locations.

To sum up, the second term in Eq.~\eqref{eq:grc} can be re-written as
\begin{equation}
\begin{aligned}
\label{eq:offset_conv}
    \sum_{n=0}^{g_h\cdot g_w-1} \sum_{c=c_n}^{c_{n+1}} \sum_{(\hat{h}, \hat{w})\in \hat{\mathcal{N}}_n} \sum_{(h,w)\in \mathcal{N}} W(c, h, w) * \\
    F(h_0+\hat{h}+h, w_0+\hat{w}+w, c)
\end{aligned}
\end{equation}
where $c_n=\frac{C}{2}+ n \cdot  \lfloor \frac{C}{2g_h\cdot g_w} \rfloor$. 

{
\textbf{Discussion on the global receptive field.} According to~\cite{luo2016understanding}, the receptive field refers to the specific region of the input data that a particular output neuron of a layer (like the convolution layer) is influenced by. In GRC, In our case, the receptive field refers to the region of $F$ covered by a unit/neuron at $F'$. Since the receptive field involves the input region and the output neuron but is not strictly related to a single convolution kernel, it can depend on the convolution channels. This observation has been largely ignored by previous work. Motivated by this, we propose the GRC to fill in the blank.

Since the output value of each unit/neuron in GRC depends on the entire region of input data, such a neuron has the global receptive field (i.e., influenced by the entire input region). The bottom row of Fig.~\ref{fig:method_sampling_loc} depicts the 2D region of the input data $F$ that influence the unit/neuron at the location $(1, 1)$ of the output feature map $F'$. It can be seen that the neuron $(1, 1)$ depends on the entire region of $F$, i.e., the red, green, purple, and yellow regions cover the whole 6$\times$6 input data. The middle row also shows the 3D version of how GRC works. We can see that when calculating the output value of neuron $(1, 1)$, different colors, each capturing different regions of the input data, can cover the entire region of the input data. In this way, we can enlarge the receptive field and learn the global dependencies.

\textbf{Further comparison with standard convolution.} 
Though the convolution operation of the standard convolution in Eq.~\eqref{eq:general_conv} and ~\eqref{eq:standard_conv} traverses all the pixels, each unit/neuron at its output feature map only depends on a limited region in the input data, preventing that unit/neuron from perceiving the global context information. For the segmentation task, if the unit/neuron at the final predicted segmentation mask (i.e., a pixel value of the mask) cannot effectively capture global context, such a unit/neuron can have a wrong class label for large objects or boundaries of objects with similar visual appearance. This is one of the major drawbacks of the standard convolution. Our GRC differs from it by leveraging the whole region of the input data to produce the outputs, owing to its channel-dependent sampling locations. Since a much larger input region is used for feature extraction, the output features of GRC can contain global context information that may not captured by the standard convolution. Thus, our GRC extracts new features with global context for better segmentation performance, compared to the standard convolution. We will also show in in Table~\ref{tab:ablation_study}, that GRC outperforms the standard convolution by a clear margin.
}

{
\textbf{Novelty and contribution of our GRC.} 
Since the core idea is relatively simple, we herein emphasize the following points to clarify its novelty and contribution.

\underline{1) Simplicity with Effectiveness}: The strength of GRC lies in its balance between conceptual simplicity and practical effectiveness. Many groundbreaking techniques in deep learning, such as residual connection~\cite{he2016deep} of ResNet and attention mechanisms~\cite{wang2018non,huang2019ccnet,zhu2019asymmetric}, stem from deceptively simple concepts. GRC aligns with this philosophy, providing a lightweight yet effective mechanism to enhance feature representation.

\underline{2) Practical Contributions}: Despite its simplicity, GRC achieves a clear improvement in performance across multiple benchmarks and applications over its baseline (see Table~\ref{tab:ablation_study}). This underscores the practical value of the proposed method, which addresses a critical need for more efficient and effective receptive field expansion in real-world tasks.

\underline{3) Unique Insights and Formulation}: Unlike traditional receptive field expansion approaches (e.g., dilated convolutions~\cite{chen2017deeplab,chen2017rethinking} or multi-scale feature aggregation~\cite{zhao2017pyramid}), GRC introduces a novel channel grouping mechanism that allows dynamic feature interaction across both local and global contexts. These formulations, e.g., Eq.~\eqref{eq:grc}-\eqref{eq:offset_conv}, differ from existing methods and add a unique dimension to receptive field manipulation.
}

{
\subsection{Theoretical Receptive Field of GRC}

To illustrate theoretical support of the global receptive field, we further analyze the theoretical receptive field of our GRC as follows.

Let us denote the receptive field at position $(h_0, w_0)$ as $R(h_0, w_0)$. For the original input images with its layer index as $l=0$, the receptive field at any position is 1 as each neuron/unit of input images is influenced by itself only. That is, for position $(h_0, w_0)$, its receptive field $R_0(h_0, w_0)=1$. For the $l$-th layer, we denote its receptive field as $R_l(h_0, w_0)$. 
Since our GRC has two groups, with the 1st group for standard convolution and the 2nd group for the global receptive field, we will first consider the receptive field of each group respectively and then discuss the overall receptive field of both groups. For the 1st group, it is standard convolution. Thus, its receptive field $R_l^1(h_0, w_0)$ is
\begin{equation}
\label{eq:receptive_1st_group}
    R_l^1(h_0, w_0) = R_{l-1}(h_0, w_0) + (K_l-1)*\prod_{p=0}^L s_p,
\end{equation}
where $K_l$ is the kernel size of layer $l$ and $s_p$ is the stride of the layer $p$. 

For the 2nd group, we have $g_h\cdot g_w$ sub-groups, each sub-group corresponding to an offset $(\hat{h}_i, \hat{w}_j), i \in [0, g_h-1], j \in [0, g_w-1]$. For the sub-group $(i,j)$, its receptive field is $R_l^2(h_0+\hat{h}_i, w_0+\hat{w}_j)$ which can be computed as per Eq.~\eqref{eq:receptive_1st_group}. Therefore, the receptive field of the 2nd group $R_l^2(h_0, w_0)$ can be obtained by enumerating all the possible combinations of $i,j$, which is
\begin{align}
    R_l^2(h_0, w_0) &= \bigcup_{i,j} R_l^2(h_0+\hat{h}_i, w_0+\hat{w}_j)\\
    &= \bigcup_{i,j} \left( R_{l-1}(h_0+\hat{h}_i, w_0+\hat{w}_j) + (K_l-1)*\prod_{p=0}^L s_p \right). \label{eq:receptive_2nd_group}
\end{align}
When enumerating all possible $(i,j)$ in the $g_h\cdot g_w$ groups, Eq.~\eqref{eq:receptive_2nd_group} can cover the entire region of the input data. This contributes to the global receptive field of our GRC. Formally, the receptive field of GRC combines the receptive fields of these two groups in Eq.~\eqref{eq:receptive_1st_group} and ~\eqref{eq:receptive_2nd_group}:
\begin{equation}
\label{eq:receptive_GRC}
    R_l^{GRC}(h_0, w_0) = R_l^1(h_0, w_0) \bigcup R_l^2(h_0, w_0).
\end{equation}
}

\subsection{Improve GRC for Efficiency}
We have elaborate on how to conduct GRC using a single filter, but note that the GRC contains $C'$ filters ($C'$ is omitted in Eq.~\eqref{eq:general_conv}). Shifting central sampling locations of all the $C'$ filters one by one, with each shifting operation being channel-dependent, is memory expensive and inefficient for parallel computation. Therefore, we further simplify the implementation of GRC to improve its efficiency by assuming that the offsets for shifting are the same for these $C'$ filters. That is, they are channel-dependent but filter-independent. Then, the simplification is based on the finding that shifting the central sampling locations of convolutional filters is equivalent to shifting the feature map because the motion is relative. 
Concretely, for each sub-group to be shifted, we keep the central sampling location the same as the standard convolution but shift the feature map of such sub-group according to the offsets. Formally, we can re-write the second term in Eq.~\eqref{eq:grc} as
\begin{align}
\label{eq:simplify_grc}
    \sum_{c=\frac{C}{2}}^{C-1}  \sum_{(h,w)\in \mathcal{N}} W(c, h, w) * 
    \hat{F}(h_0+h, w_0+w, c), \\
    \label{eq:feat_transform}
    \text{where} \quad \hat{F}(h_0, w_0, c) = F(h_0+h(c), w_0+w(c), c).
\end{align}
Eq.~\eqref{eq:simplify_grc} is standard convolution but its feature map is transformed, denoted as $\hat{F}$. $\hat{F}$ groups and then shifts the $\frac{C}{2}\sim (C-1)$-th channels of original feature map $F$ 
according to Eq.~\eqref{eq:feat_transform}, where $h(c), w(c)$ are channel-dependent shifting functions. $h(c)$ is defined as 
\begin{equation}
\label{eq:hc_wc}
    h(c) = n \cdot  \lfloor \frac{C}{2g_h\cdot g_w} \rfloor,
\end{equation}
where $n$ is channel-dependent, defined in Eq.~\eqref{eq:sub_group_index}. Similarly, we can obtain $w(c)$.

To sum up, we can first shift half channels, i.e., the $\frac{C}{2}\sim (C-1)$-th channels, of the original feature maps according to Eq.~\eqref{eq:feat_transform} \&~\eqref{eq:hc_wc}. The feature shifting will be conducted $g_h\cdot g_w$ times, much less than shifting filters (this will be $g_h\cdot g_w \cdot C'$ times). Then, we can simply apply standard convolution via Eq.~\eqref{eq:simplify_grc} to facilitate parallel computation. 
Due to fewer shifting operations and parallel computation, feature map shifting is much more efficient. In practice, we implement our GRC by shifting the feature maps.

\subsection{FCN+}
Our GRC can be used to replace the standard convolution in popular semantic segmentation networks like fully convolutional network (FCN)~\cite{long2015fully}. With this modification, we turn FCN into FCN+, which favorably improves the segmentation performance over FCN. In this section, we will detail the FCN+.

As shown in Fig.~\ref{fig:fcn_plus}, FCN+ has the same architecture as FCN, which exploits deep convolutional neural networks (CNNs) to extract features for objects and scenes. To obtain multi-scale representations, FCN+ further aggregates feature maps of different spatial resolutions from hierarchical layers. The backbone feature extractor has four stages, each stage comprising several residual blocks~\cite{he2016deep}. We apply the GRC in the last stage (Stage 5) by replacing the 2nd convolution in a residual block with GRC. Such a GRC-based block is called GRC block. The design choice of GRC block will be evaluated in experiments.

\section{Experiments}
\subsection{Experimental Setting}
\label{sec:experiment_setting}
\textbf{Datasets and protocols.}
(1) PASCAL VOC 2012~\cite{everingham2010pascal} is a popular semantic segmentation dataset. It has 21 classes (including a background class), with 1,464 images for training, 1,449 for validation, and 1,456 for testing. Most previous works~\cite{chen2017deeplab,chen2018encoder,zhang2018context,he2019adaptive,he2019dynamic} adopt its augmented training set provided by~\cite{hariharan2015hypercolumns} which comprises 10,582 training samples. Following these works, we use this augmented version as our training set.
(2) Cityscapes~\cite{cordts2016cityscapes} is a large-scale dataset for semantic understanding of urban street
scenes. It consists of 5,000 fine annotated images and 20,000 annotated images, totally 30 classes. These images are captured in 50 different cities, covering different seasons and weather conditions.
(3) ADE20K~\cite{zhou2017scene} is one of the most challenging datasets for semantic segmentation. It is challenging mainly due to its diverse and complex scenes as well as large number of classes (i.e., 150 classes). There are 20K, 2K, and 3K images for training, validation and testing respectively.

\textbf{Training details.} At the training stage, we randomly flip and scale the input images as data augmentation, with a scale ratio in $[0.5,2]$. We then set the crop size to 512$\times$512 for PASCAL VOC 2012 and ADE20K, and 512$\times$1024 for Cityscapes. Each mini-batch contains 32 cropped images. 
Following ~\cite{long2015fully}, we use ImageNet~\cite{russakovsky2015imagenet} pre-trained ResNet-101~\cite{he2016deep} as our backbone model (unless otherwise stated). We remove stride and set dilation rates 2 and 4 to the last two stages. By default, we apply the GRC in the last stage of ResNet by replacing the 2nd convolution layer in a residual block with our GRC. The hyper-parameters for the sub-group in GRC is $g_h=g_w$=4. Then, the whole model is optimized with momentum-based stochastic gradient descent (SGD). The momentum is 0.9 and the weight decay is 0.0001. The initial learning rate is 0.01 for all these datasets. The learning rate is then reduced according to different training iterations: $(1-\frac{iter}{total\_iter})^{power}$, where $power$ is 0.9. The whole training process takes 40K iterations for PASCAL VOC 2012, 80K iterations for Cityscapes, and 160K iterations for ADE20K dataset.

For inference, the testing images are also flipped and resized to multiple scales as is in the training stage. Multi-scale predictions are interpolated in a bi-linear way so that these predictions are with the same size as the input images. Then, the predictions of different scales are averaged as the final prediction.  Finally, the mean of class-wise intersection over union (mIoU) is adopted as the evaluation metric. Our experiments are conducted based on PyTorch~\cite{paszke2019pytorch} and MMSegmentation~\cite{mmseg2020}.

\subsection{Results on ADE20K}
We first conduct comprehensive ablation studies on ADE20K to show the effectiveness of our FCN+ and global receptive convolution (GRC). Then we compare the FCN+ with other state-of-the-art methods.

\textbf{GRC at different stages.} By default, we apply the GRC to the last stage of ResNet. Since there are four stages in ResNet, we further investigate which stage is the best choice. The results are shown in Table~\ref{tab:grc_stage}. It is observed that the GRC at the last stage works best, probably because the last stage contains more high-level semantic information, thus can provide better global context.

\begin{table}[t]
    \centering
    \caption{GRC vs. a variant of GRC at different stages of ResNet. The variant of GRC applies the offset-based convolution to all the channels.}
    \begin{tabular}{c|cccc}
    \hline
    \textbf{Stages} &  Stage 5 & Stage 4-5 &Stage 3-5 &Stage 2-5\\
    \hline
        GRC & 45.72 & 45.12 &45.39 &44.81 \\
        Variant of GRC  &45.52 &43.88 &42.21 &42.15 \\
    \hline
    \end{tabular}
    \label{tab:grc_stage}
\end{table}

\begin{table}[t]
    \centering
    \begin{minipage}{0.45\textwidth}
    \centering
    \caption{Comparison of GRC with deformable convolution.}
    \begin{tabular}{c|c}
    \hline
    \textbf{Methods} & mIoU \\
    \hline
        GRC & 45.72 \\
        Deformable & 44.10\\
    \hline
    \end{tabular}
    \label{tab:grc_vs_other_conv}
    \end{minipage}
~\hfill
    \begin{minipage}{0.5\textwidth}
    \centering
    \caption{GRC at different layers of a residual block.}
    \begin{tabular}{c|c}
    \hline
    \textbf{Positions} & mIoU \\
    \hline
        Conv1 (1$\times$1) & 45.31\\
        Conv2 (3$\times$3) & 45.72\\
        Conv3 (1$\times$1) & 45.39\\
    \hline
    \end{tabular}
    \label{tab:grc_layer}
    \end{minipage}
\end{table}

\begin{table}[t]
    \centering
    \caption{Sensitivity of the number of sub-groups $g$. }
    \begin{tabular}{c|cccc}
    \hline
    \textbf{Sub-groups $g$} &   4 &8 &16\\
    \hline
        mIoU & 45.72 &45.07 &45.74 \\
    \hline
    \end{tabular}
    \label{tab:sensitivity_g}
\end{table}

\textbf{GRC at different layers.} We then explore where to insert GRC in a residual block of Stage 5. A residual block of ResNet-50 or ResNet-101 has three layers, as shown in the right panel of Fig.~\ref{fig:fcn_plus}. We take turns replacing each layer with our GRC and show their performance in Table~\ref{tab:grc_layer}. It can be seen that applying GRC for the first or last 1$\times$1 convolution obtains 45.31 and 45.39 respectively, both clearly better than the baseline of FCN's 39.91 (see the 3rd row in Table~\ref{tab:ablation_study}). The results also demonstrate that the GRC at the 2nd convolutional layer works best. This can be explained by that the GRC benefits more from 3$\times$3 convolution than 1$\times$1.

\textbf{The necessity of keeping one group for standard convolution in GRC.} Recall that in Eq.~\eqref{eq:grc}, the first term is the standard convolution which aims to keep the local pixel-wise information. To justify its effectiveness, we remove the standard convolution in GRC and apply the offset-based convolution in Eq.~\eqref{eq:offset_conv} to all the channels. This modification leads to a variant of GRC as shown in the last row of Table~\ref{tab:grc_stage}. We can see that this variant of GRC decreases the performance under different stage settings. The degradation implies that (1) only utilizing the global context information is still not sufficient for accurate segmentation results; (2) the local pixel-wise information is necessary for better pixel-wise representation, which is of immense importance for dense prediction task.

\textbf{Sensitivity of the number of sub-groups in GRC.} In Eq.~\eqref{eq:spatial_group} and ~\eqref{eq:sub_group_index}, we have the two hyper-parameters for spatial-wises and channel-wise grouping, i.e., $g_h, g_w$. They determine the total number of sub-groups for channels and spatial patches. In practice, the input images are usually cropped to have the same height and width, leading to $H=W$. Thus, it is reasonable to set $g_h=g_w$ to reduce the number of hyper-parameters. Then we define $g=g_h=g_w$ and evaluate the sensitivity of $g$ in Table~\ref{tab:sensitivity_g}. We observe that the performance is stable. Though $g$ should be set according to channel depth and the size of feature maps, generally $g\in \{4,8,16\}$ is recommended.

\textbf{GRC vs. deformable convolutions.} In Table~\ref{tab:grc_vs_other_conv}, we compare our GRC with deformable convolution (standard convolution has been compared in Table~\ref{tab:ablation_study}). For fair comparison, we only replace the GRC with the deformable convolution while keep all the other settings the same. We can see from Table~\ref{tab:grc_vs_other_conv} that our GRC is clearly better than this competitor. This suggests the superiority of our GRC for enlarging the receptive field.

\begin{table}[t]
    \centering
    \caption{Comparison of FCN+ with FCN on ADE20K.}
    \begin{tabular}{c|cccc|c}
    \hline
    \textbf{Methods} &Backbone  &\#Params &FLOPs &Inf. Time &  mIoU \\
    \hline
        FCN &  ResNet-50 & 47.20M & 1.585T & 0.013s & 36.10\\
        FCN+ &  ResNet-50 & 47.20M & 1.585T &0.022s &44.54\\
    \hline
        FCN &  ResNet-101 & 66.19M & 2.208T &0.024s &39.91 \\
        FCN+ &  ResNet-101 & 66.19M & 2.208T &0.032s &45.74\\
    \hline
    \end{tabular}
    \label{tab:ablation_study}
\end{table}

\textbf{FCN+ vs. FCN.} We compare our FCN+ with FCN in Table~\ref{tab:ablation_study}. We can see that replacing the standard convolution with our proposed GRC can increase the mIoU by 8.44\% for ResNet-50, and 6.83\% for ResNet-101. It is also noteworthy that our FCN+ does not introduce any extra parameters or FLOPs to FCN. This makes our FCN+ highly efficient. The performance improvement is attributed to the global receptive field of GRC, which can exploit the global context for better pixel-wise representations. These comparisons verify the efficiency and efficacy of our FCN+.

Until now, we have investigated three cases of different channel numbers. They are A: FCN based on standard convolution with all channels having original sampling locations; B: FCN+ based on GRC with evenly separated channels; C: Variant of GRC with all channels having shifted sampling locations. Table~\ref{tab:ablation_study} shows A vs. B, where B (FCN+) is clearly better than A (FCN). Besides, Table~\ref{tab:grc_stage} shows the results of B (1st row, GRC) vs. C (2nd row) where B beats C. Thus, evenly separating channels is better than the other two design choices.

\begin{table*}[tb]
	\centering
	\caption{Segmentation results of state-of-the-art methods on PASCAL VOC 2012, Cityscapes and ADE20K. * denotes the results from the original paper with backbones differing from ResNet-101. The other results are obtained based on ResNet-101 and MMSegmentation~\cite{mmseg2020}. {The results of FCN+ are in the green cells. The best results are in bold.}}
	\begin{tabular}{p{4cm}|c|ccc}
		\hline
		\textbf{Methods}  &\textbf{\makecell[tc]{Backbones \\ (\#Params)}} &\textbf{\makecell[tc]{PASCAL \\ VOC}} & \textbf{Cityscapes} & \textbf{ADE20K}\\
		\hline
		FCN~\cite{long2015fully} & {\multirow{12}{*}{\makecell[tc]{ResNet-101 \\(42.8M)}}} & 69.91 & 75.13 &39.91\\
		PSPNet \cite{zhao2017pyramid} &  & 78.52 &79.76	&44.39\\
		PSANet \cite{zhao2018psanet} &  & 77.73 &79.31	&43.74 \\
		UperNet \cite{xiao2018unified} & & 77.43 &79.40	&43.82\\
		APCNet \cite{he2019adaptive} &  &- &79.64	&45.41 \\
		Non-local~\cite{wang2018non}  & &78.27 &78.93	&44.63\\
		DeepLabV3~\cite{chen2017rethinking} & &77.92 &80.20 &45.00\\
		DeepLabV3+~\cite{chen2018encoder} & &78.62 &\textbf{80.97}	&45.47\\
		CCNet~\cite{huang2019ccnet} & &77.87 &78.87	&43.71\\
		ANN~\cite{zhu2019asymmetric}  & & 76.70 &77.14	&42.94\\
		DNLNet~\cite{yin2020disentangled} & &- &80.41	&44.25\\ 
            DDRNet*~\cite{pan2022deep} & &- & 77.40 & -\\
		\cellcolor[HTML]{D7FFD7}\textbf{FCN+ (Ours)} &   &  \cellcolor[HTML]{D7FFD7}\textbf{79.42} &\cellcolor[HTML]{D7FFD7}80.53 &\cellcolor[HTML]{D7FFD7}45.74 \\
            \hline 
            MultiMAE*~\cite{bachmann2022multimae}  & {ViT-B (86M)} &- &- & 46.20\\
            SeMask-S FPN*~\cite{jain2023semask} & {Swin-S (50M)} &- &79.14 &\textbf{47.63} \\
		\hline

	\end{tabular}    
	\label{table:sota}
\end{table*}

\textbf{Comparison with the state of the art.} Finally, we compare our FCN+ with other state-of-the-art methods in Table~\ref{table:sota}. We can see that our FCN+ surpasses these methods based on (1) standard convolution~\cite{long2015fully,he2019dynamic}, (2) the dilated convolution~\cite{chen2018encoder,he2019adaptive}, and (3) the attention~\cite{wang2018non,huang2019ccnet,zhu2019asymmetric}. Notably, our FCN+ does not introduce any extra parameters to FCN model which is more efficient than these using multiple dilated convolutions or learning offset for each sampling location. We owe the efficacy and efficiency of FCN+ to the global receptive field brought by GRC.

{
Latest methods usually employ transformers which can also capture the long-range dependencies. For instance, MultiMAE~\cite{bachmann2022multimae} from 2022 leverages a vision transformer (ViT-Base) as the backbone to train multi-modal multi-task masked autoencoder which is then fine-tuned for semantic segmentation. SeMask~\cite{jain2023semask} from 2023 utilizes a semantic attention operation to incorporate semantic information into the Swin Transformer~\cite{liu2021swin} encoder. For these transformer-based methods, we compare their reported results using backbones with comparable parameters to our used ResNet-101 in Table~\ref{table:sota}, i.e., 86M parameters of ViT-Base\footnote{Though ViT-Base (86M) has twice parameters than ResNet-101 (42.8M), ViT-Tiny has much less parameter (21M) than ResNet-101. Thus, we use the ViT-Base for the comparison.} and 50M parameters of Swin-Small versus 42.8M parameters of ResNet-101.
Table~\ref{table:sota} shows that our FCN+ is inferior to the MultiMAE and SeMask on ADE20K. Specifically, FCN+ is 0.46 worse than MultiMAE but its backbone has only half the parameters of the ViT-B, i.e., 42.8M versus 86M. Therefore, we argue that the performance of FCN+ is comparable to the ViT-based method. Compared with the latest Swin Transformer-based method, FCN+ shows a 1.89 gap to SeMask. This is mainly because the Swin Transformer can also capture the global context using the attention mechanism. However, we will see that FCN+ can outperform Swin Transformer-based SeMask on Cityscapes despite the failure on ADE20K. As such, we believe that the GRC-based network can still excel in some scenarios compared with the transformer.
}

\subsection{Results on Cityscapes}
Table~\ref{table:sota} also shows the comparison on Cityscapes. 
{
Except for the methods compared on ADE20K, we further compare with DDRNet~\cite{pan2022deep} from 2022 which also seeks to enlarge the receptive field of convolution. 
}
It is clear that our FCN+ improves the FCN by 5.40\%, and achieves the second best performance among all the competitors. 
{
Notably, FCN+ can achieve better performance than DDRNet and Swin Transformer-based SeMask. This means that our GRC can effectively improve the performance of convolutional neural networks like FCN to beat the latest transformers, which illustrates the effectiveness of the GRC-based network over the transformer-based methods in some scenarios. 
Our GRC and FCN+ only use a simple decoder of FCN head~\cite{mmseg2020},
}
this verifies that even without complicated decoders, the FCN+ can effectively deal with the challenges of objects' scale variation and confusion boundaries.

\subsection{Results on PASCAL VOC 2012}
We further evaluate our FCN+ on PASCAL VOC 2012 and compare it with other state-of-the-art methods in Table~\ref{table:sota}. FCN+ increases the performance of FCN by 9.51\%, which demonstrates the superiority of GRC to the standard convolution. Compared with other methods, the FCN+ also obtains comparable or even better performance. The superior performance of FCN+ illustrates that GRC can handle the challenge of complex scenes and diverse classes in PASCAL VOC 2012.

{
\subsection{Cross-dataset Evaluation for Unsupervised Domain Adaptation Tasks}

To validate the generalization ability of our FCN+, we conduct an Unsupervised Domain Adaptation (UDA) task that trains the model on GTA5~\cite{richter2016playing} and tests it on Cityscapes. GTA5 dataset has 24,966 synthesized images extracted from the photo-realistic computer game of GTA5. Its class labels are compatible with Cityscapes, thus facilitating cross-dataset evaluation. 
To complete the UDA task, GTA5 is used as the labeled source domain while Cityscapes is adopted as the unlabeled target domain. The training sets of both datasets are then utilized to train the model, with the labels of the target dataset (i.e., Cityscapes) discarded. The validation set of Cityscapes is used as the test set, which follows the setting in ~\cite{saito2018maximum}. 

We use ResNet-101-based FCN and FCN+ as two different backbones for a well-known UDA method, Maximum Classifier Discrepancy (MCD)~\cite{saito2018maximum}. For FCN+ containing GRC blocks in Figure~\ref{fig:fcn_plus}, its setting is consistent with the default one stated in Section~\ref{sec:experiment_setting} where the sub-group in GRC is $g_h=g_w$=4. 
Following ~\cite{saito2018maximum}, we train them for 10 epochs using Momentum SGD, with a momentum of 0.9. The learning rate is 0.001 and the batch size is 1. The images are resized to 512$\times$1024 for training. After training, we report the mIoU of both models after 10 epochs on the test set (i.e., the validation set of Cityscapes). 

The results in Table~\ref{tab:uda_gta_city} show that FCN+ outperforms FCN by 1.22\% on the cross-dataset evaluation task. The superiority of FCN+ demonstrates that GRC can be better generalized across semantic segmentation datasets than the standard convolution in FCN.

}

\begin{table}[tb]
    \centering
    \caption{{Unsupervised domain adaptation results of MCD~\cite{saito2018maximum} with FCN and FCN+ as backbones, respectively. The mIoU of adaptation from GTA5 to Cityscapes is reported.}}
    \begin{tabular}{c|c|c}
    \hline
      \textbf{Method}   &  MCD (FCN) & MCD (\textbf{FCN+})\\
    \hline
      \textbf{mIoU}   & 23.73 & 24.95\\
    \hline
    \end{tabular}    
    \label{tab:uda_gta_city}
\end{table}

\subsection{Generalization to Image Classification Tasks}
Though our GRC is proposed for semantic segmentation tasks, its capability of capturing global and local information can possibly contribute to image classification tasks. To investigate the GRC on image classification tasks, we apply the GRC to classify 100 categories in CIFAR-100~\cite{krizhevsky2009learning}. CIFAR-100 has 50,000 images for training and 10,000 for testing, each with a size 32$\times$32. We use Wide ResNet-28-2~\cite{zagoruyko2016wide} as the backbone and optimize it using an SGD optimizer for 100 epochs. The initial learning rate is 0.05 and is updated using a cosine learning decay policy~\cite{loshchilov2016sgdr}. The first standard convolution in the first block of the backbone is replaced with our GRC of which the group number $g_h=g_w=4$. The results are reported in Table~\ref{tab:img_classification}.
\begin{table}[htb]
    \centering
    \caption{The accuracy on CIFAR-100.}
    \begin{tabular}{c|c|c}
    \hline
     \textbf{Backbones} & Wide ResNet-28-2 & Wide ResNet-28-2 + GRC\\
    \hline
       \textbf{Accuracy}  & 64.21\% & 65.06\% \\
    \hline
    \end{tabular}
    \label{tab:img_classification}
\end{table}

We can see that incorporating GRC can improve the performance by 0.85\%, implying that GRC can also be generalized to image classification tasks to a certain extent. We also notice that the gap may not be as significant as in semantic segmentation, probably because the small input image size of CIFAR-100 (i.e., 32$\times$32) does not contain large-scale objects, making the global receptive field less effective. Despite this, our GRC can still work better than the standard convolution, suggesting the strong generalization ability of GRC to various tasks other than semantic segmentation.

\subsection{Limitations and Social Impacts}
\label{sec:limitations}
Though our GRC is efficient and effective, it relies on artificial grouping and introduces a hyper-parameter $g(=g_w=g_h)$. These groups can sometimes be sub-optimal, e.g., a specific large object can be grouped into several sub-groups. This can possibly be alleviated by designing content-dependent grouping--- automatically grouping according to content can be more effective and save time for hyper-parameter tuning.

Another limitation of GRC is that it may take more inference time and slow down the testing speed. This is because the shifting operation is not memory friendly.  We assume that optimizing the implementation of GRC to take advantage of CUDA can further speed up inference, like code optimization for CUDA in deformable convolution.

One of the negative social impacts of our work is that our method may suffer from the safety issue caused by adversarial attacks. To ensure the safety of applying our work to real-world applications, further analysis on the robustness of our method to adversarial attacks is planned to be conducted.

\section{Conclusion}
In this paper, we find that a size-fixed filter cannot capture global receptive field mainly because the grid sampling locations of such a filter are limited to a fixed local spatial coordinate. To address this issue, we argue that the grid sampling locations of convolution should be dependent on both spatial coordinates and different channels. Based on this, we propose a novel global receptive convolution (GRC) to provide global receptive field for convolution. More importantly, the GRC can integrate the global context into the original location information of each pixel for better dense prediction results. We further incorporate the GRC into FCN and propose FCN+. We show that FCN+ can outperform state-of-the-art methods on some popular semantic segmentation datasets, such as PASCAL VOC 2012, Cityscapes and ADE20K.

\section*{Acknowledgment}
This work was supported by the National Key R\&D Program of China (Grant No. 2023YFB3907405), the Chinese Academy of Sciences Project for Young Scientists in Basic Research (YSBR-037). This project was also supported by the National Science Foundation for Young Scientists of China (Grant No. 42105113).

\bibliographystyle{elsarticle-num}
\bibliography{ref.bib}

\end{document}